\pdfoutput=1

\documentclass[11pt]{article}

\usepackage[]{latex/acl}

\usepackage{times}
\usepackage{latexsym}

\usepackage[T1]{fontenc}

\usepackage[utf8]{inputenc}

\usepackage{microtype}

\usepackage{inconsolata}

\usepackage{graphicx}
\usepackage{algorithm}
\usepackage{algorithmic}
\usepackage{threeparttable}
\usepackage{multirow}
\usepackage{booktabs}
\usepackage{amssymb}
\usepackage{subfig}
\usepackage{amsmath}
\usepackage{url}
\usepackage{fontawesome}

\definecolor{brandeisblue}{rgb}{0.0, 0.44, 1.0}
\definecolor{carrotorange}{rgb}{0.93, 0.57, 0.13}
\definecolor{x11gray}{rgb}{0.75, 0.75, 0.75}

\newcommand{\ignore}[1]{}
\title{Learning High-Quality and General-Purpose Phrase Representations }

\author{
Lihu Chen\textsuperscript{\rm 1},
Gaël Varoquaux\textsuperscript{\rm 1}, 
Fabian M. Suchanek\textsuperscript{\rm 2} \\
\textsuperscript{\rm 1} Soda, Inria Saclay, France \\
\textsuperscript{\rm 2} LTCI, Télécom Paris, Institut Polytechnique de Paris, France \\
\texttt{\{lihu.chen, gael.varoquaux\}@inria.fr}\\ 
\texttt{fabian.suchanek@telecom-paris.fr}
}

\begin{document}
\maketitle
\begin{abstract}
Phrase representations play an important role in data science and natural language processing, benefiting various tasks like Entity Alignment, Record Linkage, Fuzzy Joins, and Paraphrase Classification.
The current state-of-the-art method involves fine-tuning pre-trained language models for phrasal embeddings using contrastive learning. However, we have identified areas for improvement. First, these pre-trained models tend to be unnecessarily complex and require to be pre-trained on a corpus with context sentences.
Second, leveraging the phrase type and morphology gives phrase representations that are both more precise and more flexible.
We propose an improved framework to learn phrase representations in a context-free fashion.
The framework employs phrase type classification as an auxiliary task and incorporates character-level information more effectively into the phrase representation.
Furthermore,  we design three granularities of data augmentation to increase the diversity of training samples.
Our experiments across a wide range of tasks show that our approach generates superior phrase embeddings compared to previous methods while requiring a smaller model size. \\
\faGithub~ Code: \url{https://github.com/tigerchen52/PEARL}\\
\faLink~ PEARL-small: \url{https://huggingface.co/Lihuchen/pearl_small}\\
\faLink~ PEARL-base: \url{https://huggingface.co/Lihuchen/pearl_base}
\end{abstract}.

\section{Introduction}
A phrase is a group of words (or a single word) with a special meaning. They may denote recognizable entities: names of people (\textit{Albert Einstein}), organizations (\textit{The New York Times}), dates (\textit{23 February 2008}), and events (\textit{2024 Summer Olympics}). 
Beyond these typical contexts, phrases also appear as column names in tabular data (\textit{average\_wage}), as user queries (\textit{black pant men}), or even as a non-noun phrase in clinical reports (\textit{more than 63kg}).
Phrases are thus an important building block in many applications of both data science and natural language processing (NLP), e.g., in tasks such as Entity Alignment~\cite{zhao2020experimental}, Fuzzy Joins~\cite{yu2016string}, Question Answering~\cite{lee2021learning}, Record Linkage~\cite{christen2011survey}, and Syntactic Parsing~\cite{socher2010learning}.
Central to these applications is 
the assessment of the semantic similarity between two distinct phrases. Today, the main tool to assess the similarity of phrases is \emph{phrase embeddings}, i.e., learned vector representations that capture the semantics of the phrases in such a way that phrases with similar meanings are close in representation space. 

\begin{figure}[t]
	\centering
	\includegraphics[width=0.5\textwidth]{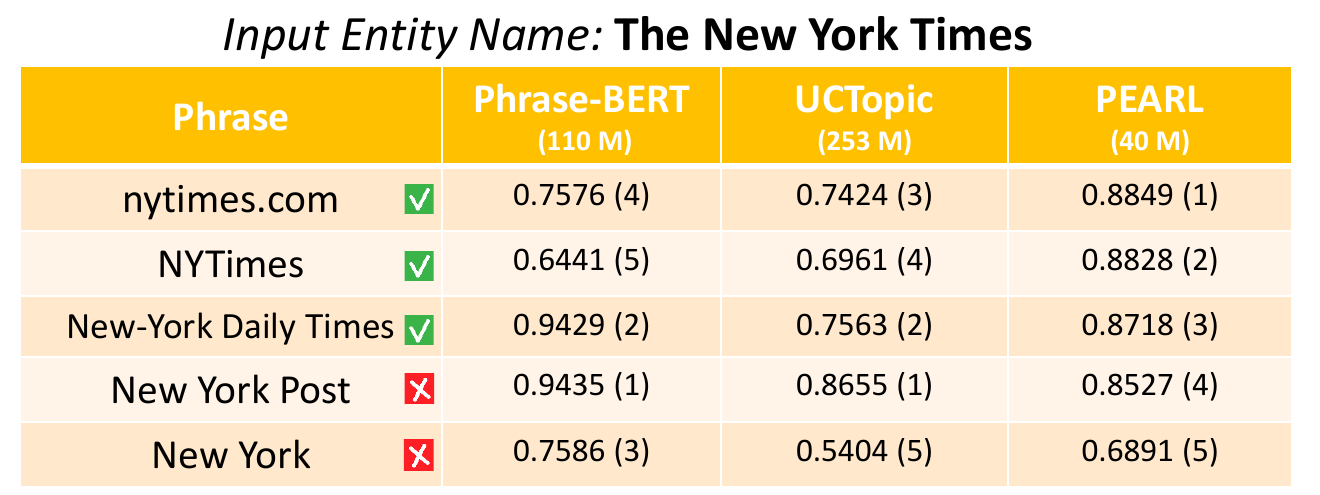}
	\caption{An example of entity retrieval. Given the input entity name \textit{``The New York Times''}, we show the cosine similarity obtained by different models. The ranking of scores is listed in parentheses. }
	\label{fig:intro_example}
\end{figure}

The difficulty of learning such representations arises from the fact that phrases often appear without context (e.g., in user queries),  and exhibit diverse morphological variations. 
For example, given the entity \textit{``The New York Times (Q9684)''}, the knowledge base Wikidata~\cite{vrandevcic2014wikidata} offers multiple aliases (alternate names)\footnote{\url{https://www.wikidata.org/wiki/Q9684}}. Three of them are shown in the first rows of Figure~\ref{fig:intro_example}. The last two rows show names of other entities: \textit{``New York Post (Q211374)''} and \textit{``New York (Q1384)''}.
While all five of these phrases look very much alike, only the first three are associated with \textit{``The New York Times''}.
This versatility of phrases makes it hard to use rule-based or string-distance methods for semantic similarities.
Sentence-BERT~\cite{reimers2019sentence} was the first approach to fine-tune pre-trained language models like BERT~\cite{devlin2018bert} and RoBERTa~\cite{liu2019roberta} to derive meaningful sentence embeddings. 
However, Sentence-BERT is given entire sentences during training (no special focus on 
short texts or phrases), so that its capabilities to embed phrases remain limited.
Phrase-BERT~\cite{wang2021phrase} was explicitly designed to embed phrases and adopts contrastive learning to fine-tune BERT on lexically diverse phrasal paraphrase pairs and their surrounding context, yielding more powerful phrase embeddings.
Another context-aware approach, UCTopic~\cite{li2022uctopic}, further improved phrase representations by using cluster-assisted negative sampling i.e., leveraging clustering results as pseudo-labels.

\begin{figure*}[t]
	\centering
	\includegraphics[width=1.0\textwidth]{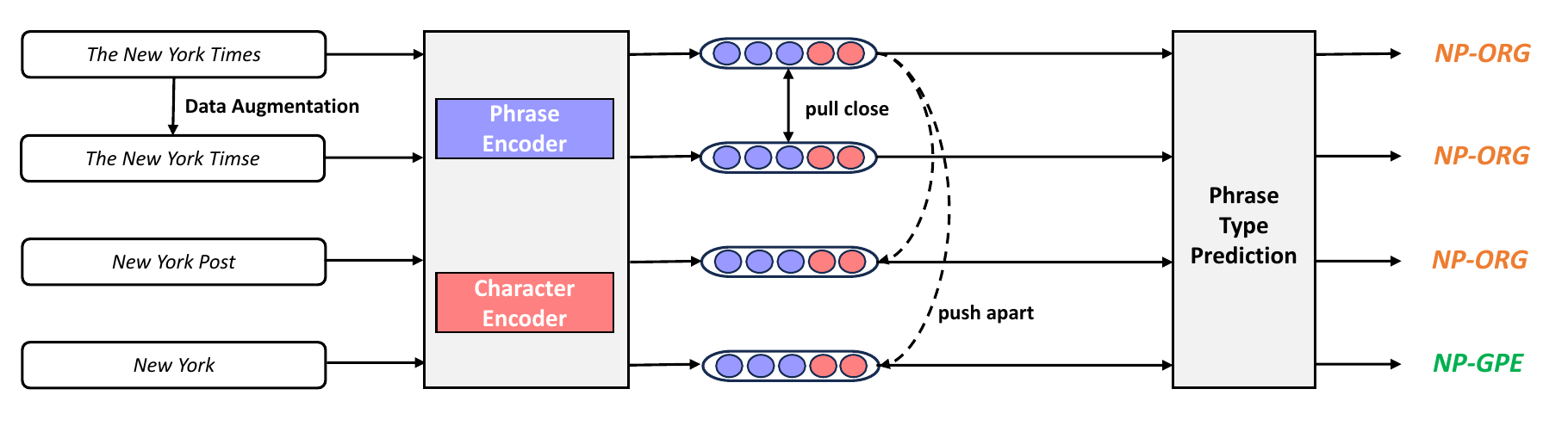}
	\caption{An illustration of PEARL. It uses contrastive learning and an auxiliary task of phrase type prediction for learning phrase embeddings.  }
	\label{fig:framework}
\end{figure*}

However, this prior work faces several limitations. 
First,  phrases frequently appear devoid of context cues, especially in tabular data, and are often characterized by short lengths. Consequently, we might not actually need the complex reasoning abilities of large (or deep) language models.
A small (or shallow) neural architecture could suffice for the purpose of capturing phrase semantics. Also, we need a model that works well in the absence of context.
Second, existing work partially neglects the type information of phrases. For example, 
although \textit{``The New York Times''} and \textit{``New York''} have a high lexical overlap, a good representation model should distinguish them since the first phrase pertains to an organization while the second is linked to a geopolitical entity. 
Third, existing sub-word embeddings are not robust against out-of-vocabulary words~\cite{chen2022imputing}, and this vulnerability entails the necessity of using character-level features and morphological information. Indeed, as Figure~\ref{fig:intro_example} shows, Phrase-BERT and UCTopic fail to recognize that \textit{``NYTimes''} is an abbreviation of the original phrase, and wrongly rank \textit{``New York Post''} (a different newspaper) as closest to \textit{``The New York Times''}. 

In this paper, we present a context-free contrastive learning framework called PEARL\footnote{Phrase Embeddings by Augmented Representation Learning}, which enriches existing language models by incorporating phrase type and character-level features. 
Additionally, PEARL uses a range of data augmentation techniques to increase training samples.
PEARL has the following advantages: 
First, it is able to discern between phrases that share similar surface forms but are of different semantic types. 
For example, a model using our framework sees \textit{``New York''} as a poor match for \textit{``The New York Times''} as it is of a different type: a geopolitical entity versus an organization (Figure~\ref{fig:intro_example}).
Second, our approach captures morphology in phrases better. In Figure~\ref{fig:intro_example}, 
our method correctly ranks all three positive candidates, including those with acronyms, as \textit{NYTimes}.
Third, a PEARL model of relatively small size (40M parameters) can outperform existing larger models (Phrase-BERT and UCTopic) and it learns phrase embeddings in a context-free fashion. 
This results in shorter training times and less resource consumption, which makes our approach more accessible in low-resource scenarios and reduces its carbon footprint.

We conduct extensive experiments with PEARL across various phrase and short text tasks, including Paraphrase Classification, Phrase Similarity, Entity Retrieval, Entity Clustering, Fuzzy Join, and Short Text Classification.
We can show that our method outperforms other competitors across all these tasks -- despite a smaller model size.

\section{Related Work}
Phrases are fundamental linguistic units, pivotal to understanding languages. Hence, learning their representations has attracted quite some attention in the research community.  
Early works mostly use compositional transformation to obtain phrasal embeddings, i.e., they derive phrase representations from word embddings~\cite{mitchell2008vector,socher2012semantic, hermann2013role, yu2015learning, zhou2017learning}.  With the advent of large pre-trained models, recent approaches fine-tune transformer models like BERT~\cite{devlin2018bert} to obtain generalized text embeddings, e.g. Sentence-Bert~\cite{reimers2019sentence} and E5~\cite{wang2022text}.
However, a recent study suggests that phrase representations in these language models heavily rely
on lexical content while struggling to capture the sophisticated compositional semantics~\cite{yu2020assessing}. 
To develop more powerful models dedicated to phrasal representations, Phrase-BERT~\cite{wang2021phrase} fine-tunes BERT on lexically
diverse datasets by using both phrase-level paraphrases and context sentences around phrases. This allows the production of embeddings that go beyond simple lexical overlap.
Another context-aware model, UCTopic~\cite{li2022uctopic}, proposes cluster-assisted contrastive learning for inducing phrasal representations for topic mining. 
McPhraSy~\cite{cohen2022mcphrasy} incorporates context information into phrase embeddings during inference. 
Although these methods can effectively generate semantically meaningful phrasal representations, they ignore the phrase type and morphological information, which are crucial for understanding phrases. 
In this paper, we show that our approach can outperform these models with a much smaller model.

In the field of data science, a task closely related to phrase representation is string matching. It is widely used across diverse applications, including Fuzzy Join~\cite{yu2016string}, Entity Resolution~\cite{papadakis2020blocking} or Alignment~\cite{zhao2020experimental}, and Ontology Matching~\cite{otero2015ontology}.
A simple yet effective solution for this task is similarity functions such as the Edit Distance and Jaccard similarity, which assess either token-level or character-level (or n-gram) similarity.
More refined methods resort to word embeddings like GloVe~\cite{pennington2014glove} and Fasttext~\cite{bojanowski2017enriching} to better capture lexical meaning.
In this work, we show that models trained by our framework can be used for a series of database or knowledge base related tasks and achieve competitive results at little cost.

\section{Our Approach}
Our objective is to learn representations for arbitrary input phrases. For this, we design a novel contrastive-learning framework named PEARL, as shown in Figure~\ref{fig:framework}. 
The input for PEARL is \textbf{\emph{context-free phrases}}. This is different from other existing models like Phrase-BERT~\cite{wang2021phrase} and UCTopic~\cite{li2022uctopic} which take phrases with context as input.
Given a specific phrase, PEARL first applies data augmentation in order to obtain similar phrases that will serve as positive samples.
For example, \textit{``The New York Times''} becomes \textit{``The New York Timse''} by using a character-level augmentation (character swap). 
Next, embeddings are generated by both phrase-level and character-level encoders. 
We then learn embeddings with the help of contrastive loss, which aims to pull close positive pairs while pushing apart in-batch negative samples.
In order to learn more expressive representations, we add a certain number of hard negatives to each batch. For example, 
\textit{``New York Post''} and \textit{``New York''} can serve as hard negatives, given their high lexical overlap with the original phrase coupled with very distinct semantics. 
To integrate phrase structural information into the representations, we
force the framework to assign tags of a lexical class and a named entity type to each phrase. 
For example, the framework learns to assign a \texttt{NP-ORG} tag to the phrase \textit{``The New York Times''}, meaning that the phrase is a noun phrase associated with an organization. The negative sample \textit{``New York''}, in contrast, receives a \texttt{NP-GPE} tag, meaning that the phrase is a noun phrase linked to a geopolitical entity. 
This augmentation with entity type information allows the model to distinguish \textit{``The New York Times''} and \textit{``New York''} 
in the representation space.


\begin{figure}[t]
	\centering
	\includegraphics[width=0.5\textwidth]{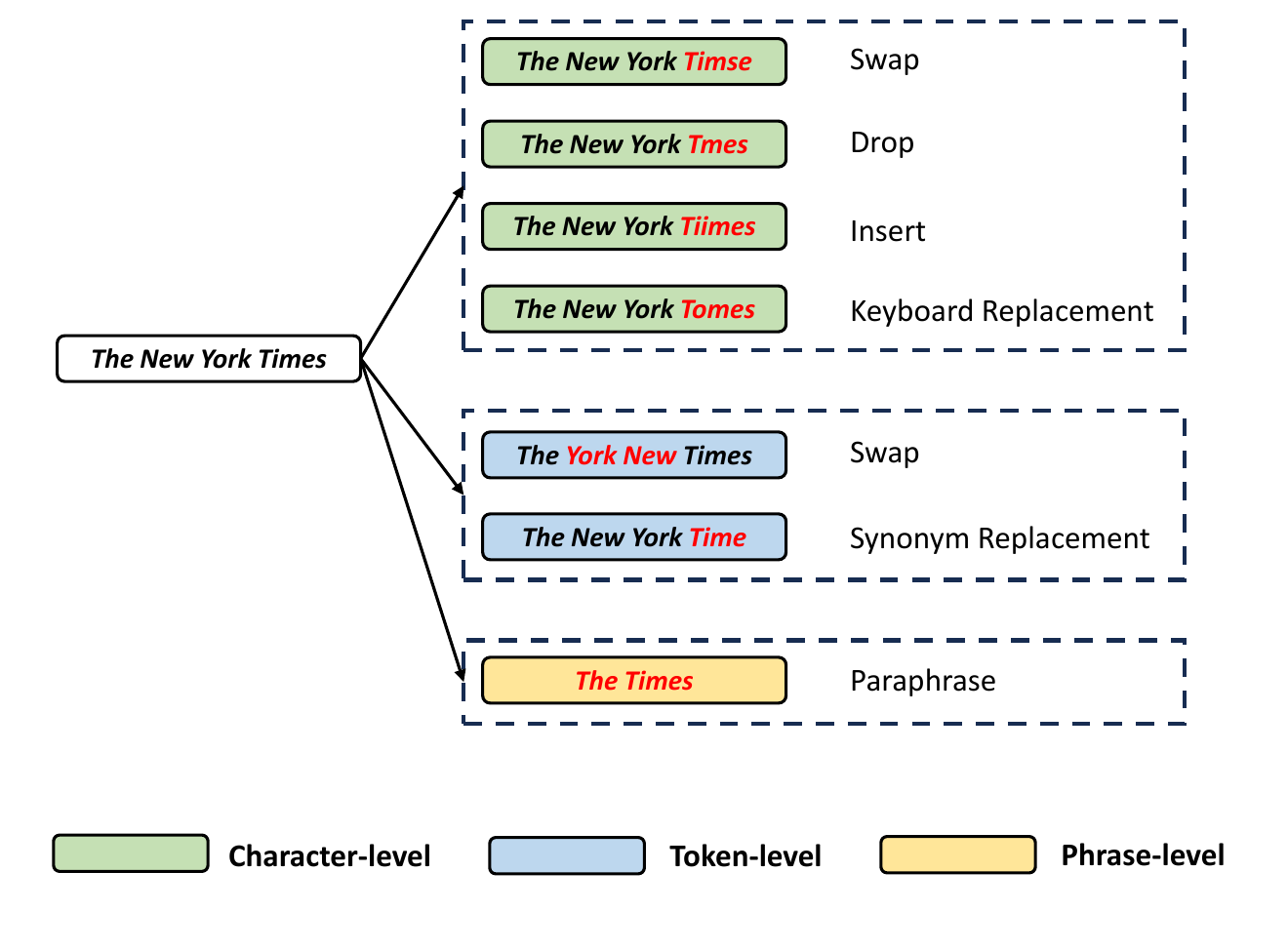}
	\caption{Different levels of granularity for the data augmentation methods on \textit{``The New York Times''}.}
	\label{fig:data_augmentation}
\end{figure}

\subsection{Data Augmentation}
The positive pairs in contrastive learning are generated by data augmentation, and we use three different granularity methods to create training samples, as shown in Figure~\ref{fig:data_augmentation}. 

\paragraph{Character-level Augmentation}$\!\!\!$aims to add morphological perturbations to the characters inside a single word. The goal is to make the representations robust against variations so that phrases that have the same meaning but slightly different surface forms (e.g., misspellings) can be pulled close in the representation space.  
We adopt four types of character-level augmentations, inspired by Out-of-Vocabulary models~\cite{pruthi2019combating,chen2022imputing}: (1) Swap two consecutive characters, (2) drop a character, (3) insert a new character, (4) replace a character according to keyboard distance.

\paragraph{Token-level Augmentation}$\!\!\!$modifies tokens in phrases for constructing positive samples.
One method is to swap the order of two adjacent tokens, 
as in \textit{``New York''} $\rightarrow$ \textit{``York New''}. Another method is Synonym Replacement, which substitutes a token in a phrase with a synonymous one from a lexical dictionary.
For example, \textit{``New York newspaper''} can be transformed to \textit{``NYC newspaper''}.
We use two methods to retrieve synonyms: First, we draw synonyms from the lexical database WordNet~\cite{miller1995wordnet}.
Second, we use the word embeddings of FastText~\cite{bojanowski2017enriching}. We regard word pairs whose vector cosine similarity is greater than a certain threshold as synonyms.  

\paragraph{Phrase-level Augmentation}$\!\!\!$paraphrases an input phrase for generating completely diverse samples. 
Specifically, we employ a text-to-text paraphraser called Parrot~\cite{prithivida2021parrot}.
For instance, consider the input phrase \textit{``The New York Times''}.
Through the usage of Parrot, an alternative name such as \textit{``The Times''}\footnote{\textit{``The Times''} is an ambiguous name, and it can also mean a British daily national newspaper based in London. } can be generated as output. 
This augmentation stands distinct from character and token methods, thereby broadening the diversity of positive samples.

\subsection{Encoder}
Phrases that are semantically similar can differ both on the token level (as in  \textit{``adult male''} vs. \textit{``grown man''}) and on the character level (as in \emph{adult} vs. its typo \emph{adlut}). To cater to both variations, 
we feed the input phrase into both a phrase-level encoder and a character-level encoder and concatenate the two embeddings. 

\paragraph{Phrase Encoder.} We use E5~\cite{wang2022text} as our phrase encoder. 
E5 is a general-purpose text embedding model pre-trained on curated large-scale (270 million) text pairs.
It is able to transfer to a wide range of tasks requiring a single-vector representation of texts such as classification, retrieval, and clustering.

\paragraph{Character Encoder.} We take inspiration from LOVE~\cite{chen2022imputing}, a lightweight out-of-vocabulary model, to generate character-level embeddings.
LOVE can produce word embeddings for arbitrary unseen words such as misspelled words, rare words, and domain-specific words, and it learns the behavior of pre-trained embeddings using only the surface form of words. 
We feed the vector obtained by LOVE to a fully connected layer to reduce its dimension.

\subsection{Phrase Type Classification}
The semantic type of a phrase is an important piece of information for distinguishing phrases that share similar surface forms but possess different meanings (such as \textit{``The New York Times''} and \textit{``New York''}).
To integrate the phrase type into the learning framework, we design an auxiliary training task, Phrase Type Classification, which aims to predict the tags of the lexical phrase class and entity types for an input phrase.  
We use the following lexical tags during training: Noun Phrase (\texttt{NP}), Verb Phrase (\texttt{VP}), Prepositional Phrase (\texttt{PP}), Adverb Phrase (\texttt{ADVP}), and Adjective Phrase (\texttt{ADJP}).
As for the entity type, we use the named entity labels defined in OntoNotes~\cite{hovy2006ontonotes}: \texttt{CARDINAL}, \texttt{DATE}, \texttt{PERSON}, \texttt{NORP}, \texttt{GPE}, \texttt{LAW}, \texttt{PERCENT}, \texttt{ORDINAL}, \texttt{MONEY}, \texttt{WORK\_OF\_ART}, \texttt{FAC}, \texttt{TIME}, \texttt{QUANTITY}, \texttt{PRODUCT}, \texttt{LANGUAGE}, \texttt{ORG}, \texttt{LOC}, and \texttt{EVENT}. 
We add an \texttt{OTHER} for phrases that do not belong to any of them.
We combine the two sets in a Cartesian product so that we obtain a label set $\mathcal{Y}$ with 95 phrase types in total.
For example, the label \texttt{NP-GPE} signifies a noun phrase related to a geopolitical name (\textit{``the United States''}), a label  \texttt{VP-ORG} corresponds to a verb phrase associated with an organization (\textit{``Bring Me the Horizon''}), and a label \texttt{PP-QUANTITY} identifies a propositional phrase linked to a quantity (\textit{``between 1500 to 2000 ft''}), which might be useful for numerical reasoning tasks.

Now suppose that we have an $m$-dimensional vector $\mathbf{u} \in \mathbb{R}^{m}$ and an $n$-dimensional vector $\mathbf{v} \in \mathbb{R}^{n}$  generated by the phrase and character encoder, respectively. We concatenate them and apply a softmax layer with a trainable weight $\mathbf{W} \in \mathbb{R}^{(m+n) \times |\mathcal{Y}|}$:
\begin{equation}
\mathbf{o}^{et} = \mathrm{softmax}((\mathbf{u, v}) \mathbf{W})
\end{equation}
\noindent Here, $\mathcal{Y}$ is the label set and  $\mathbf{o}^{et} \in \mathbb{R}^{|\mathcal{Y}| }$ is the final output for predicting the entity type.

\subsection{Objective and Training}

\paragraph{Loss Function.}
There are two training tasks in our framework: Contrastive Learning and Phrase Type Classification.
We adopt the widely-used contrastive loss~\cite{hjelm2018learning, chen2020simple} for training, which 
encourages learned representations for positive pairs to be close while pushing apart representations of negative samples. 
The loss function can be written as:
\begin{align}\label{eq:cl_loss}
	\mathcal{L}_{\text{CL}} &= - \log \frac{e^{\text{sim}(\mathbf{h}^{\mathsf{T}}\mathbf{h}^{+})/ \tau} }{ e^{\text{sim}(\mathbf{h}^{\mathsf{T}}\mathbf{h}^{+})/ \tau} + \sum e^{\text{sim}(\mathbf{h}^{\mathsf{T}}\mathbf{h}_{i}^{-})/ \tau}}
\end{align}
\noindent Here, $\tau$ is a temperature parameter that regulates the level of attention given to difficult samples, $\text{sim}(\cdot)$ is a similarity function such as cosine similarity, and $(\mathbf{h},\mathbf{h}^{+})$, $(\mathbf{h},\mathbf{h}^{-})$ are positive pairs and negative pairs, respectively (assuming that all vectors are normalized).
During training, we apply one data augmentation randomly to the original phrase for obtaining positive pairs while negative examples are the other samples in the mini-batch.
This training process encourages the model to learn representations that are invariant against variations.

As for the task of Phrase Type Classification, we use a standard cross-entropy loss:

\begin{equation}
\mathcal{L}_{\text{CE}} = -{\textstyle \sum_{i=1}^{|\mathcal{Y}|}} y_{i}\log o^{et}_{i}
\end{equation} 
\noindent Finally, the overall learning objective is:
\begin{equation}
\mathcal{L} = \mathcal{L}_{\text{CL}} + \mathcal{L}_{\text{CE}}
\end{equation} 

\paragraph{Training Corpus.}
We use Wikipedia to construct our training samples. We parse the articles with the Berkeley Neural Parser~\cite{kitaev2018constituency} and collect five lexical types of phrases (\texttt{NP, VP, PP, ADVP, ADJP}). We remove phrases that appear less than two times and obtain around 3.8 million phrases in total (\texttt{NP}: 60.1\%, \texttt{VP}: 0.4\%, \texttt{PP}: 26.1\%, \texttt{ADVP}: 11.0\%,  \texttt{ADJP}: 2.4\%).
To obtain the entity types, we employ a Named Entity Recognition (NER) model. 
We use DeBERTa~\cite{he2020deberta} fine-tuned on OntoNotes~\cite{hovy2006ontonotes}.
The entity type distribution is shown in Figure~\ref{fig:phrase_type_dis}.

\paragraph{Hard Negatives.}
Conventional contrastive learning regards other samples in the same batch as negatives (in-batch negatives)~\cite{hjelm2018learning, chen2020simple}, which is simple and effective.
However, these negative samples might be easy to distinguish by a model. For example, 
\textit{``The New York Times''} and \textit{``two years after''} can be in the same batch during training, but this negative pair contributes less to the parameter optimization process.
Hence, we introduce \emph{hard negatives} into each batch, i.e., samples that have a surface form similar to the original phrase, but a different semantics -- as in \textit{``The New York Times''} and \textit{``New York City''}. 
For each phrase in the training set, we first retrieve candidates that have a small edit distance with the original phrase. Next, all the candidates are encoded by the E5 text embedding. Finally, the candidates with a low cosine similarity are selected as the hard negatives.
During training, a certain number of hard negatives are added to each batch. 

\paragraph{Weight Average.} We found that there is a catastrophic forgetting problem~\cite{mccloskey1989catastrophic} after fine-tuning, i.e., the model forgets previously learned information upon learning new information. 
To avoid this, we average the weights of the original and fine-tuned models, which is simple yet effective.

\section{Experiments}

\subsection{Datasets}
To evaluate our framework, we use tasks of phrase and short text in experiments.
In total, there are six types of tasks, 
which cover both the field of data science and of natural language processing. 
We briefly introduce tasks and datasets used in experiments and you can see more details in the appendix~\ref{sec:details_of_datasets}.

For phrase datasets, we consider five tasks:
(1) \textbf{Paraphrase Classification}.  We use two paraphrase classification datasets used by Phrase-BERT~\cite{wang2021phrase}: \emph{PPDB} and \emph{PPDB-filtered}. 
(2) \textbf{Phrase Similarity}. We use two datasets, \emph{Turney}~\cite{turney2012domain} and \emph{BIRD}~\cite{asaadi2019big}.
(3) \textbf{Entity Retrieval}. We construct two entity retrieval datasets by using a general knowledge base \emph{Yago}~\cite{pellissier2020yago} and a biomedical terminology \emph{UMLS}~\cite{bodenreider2004unified}, respectively. 
(4) \textbf{Entity Clustering}. We use the general-purpose \emph{CoNLL~03}~\cite{sang2003introduction} benchmark and the biomedical \emph{BC5CDR}~\cite{li2016biocreative} benchmark.
(5) \textbf{Fuzzy Join}.  We use the \emph{AutoFJ} benchmark~\cite{li2021auto}, which contains 50 diverse fuzzy-join datasets derived from DBpedia~\cite{lehmann2015dbpedia}.

\begin{table*}[bt] 
	\centering 
	\scriptsize
	\setlength{\tabcolsep}{0.9mm}{
		\begin{threeparttable} 
			\begin{tabular}{cccccccccccccc}  
				\toprule
				\textbf{Model}
				&\textbf{Size}&\multicolumn{2}{c}{\textbf{\underline{Paraphrase Classification}}}&\multicolumn{2}{c}{\textbf{\underline{Phrase Similarity}}}&\multicolumn{2}{c}{\textbf{\underline{Entity Retrieval}}}&\multicolumn{2}{c}{\textbf{\underline{Entity Clustering}}} &\multicolumn{1}{c}{\textbf{\underline{Fuzzy Join}}} &Avg\cr
				&&PPDB&PPDB filtered&Turney&BIRD&YAGO&UMLS&CoNLL 03&BC5CDR&AutoFJ& \cr
				Length&&(2.5)&(2.0)&(1.2)&(1.7)&(3.3)&(4.1)&(1.5)&(1.4)&(3.8)&(2.4) \cr
				\midrule
				String Distance &-&-&-&-&-&-&-&-&-&64.7&-\cr
				GloVe~\shortcite{pennington2014glove} &-&95.5&50.6&31.5&53.1&20.9&18.8&21.2&7.8&50.6&38.9\cr
				FastText~\shortcite{bojanowski2017enriching} &-&94.4&61.2&59.6&58.9&16.9&14.5&3.0&0.2&53.6&40.3\cr
				Sentence-BERT~\shortcite{reimers2019sentence} &110M&94.6&66.8&50.4&62.6&21.6&23.6&25.5&48.4&57.2&50.1\cr
				Phrase-BERT~\shortcite{wang2021phrase} &110M&96.8&68.7&57.2&68.8&23.7&26.1&35.4&59.5&66.9&54.5\cr
				UCTopic~\shortcite{li2022uctopic} &240M&91.2&64.6&\underline{60.2}&60.2&5.2&6.9&18.3&33.3&29.5&41.6\cr
				E5-small~\shortcite{wang2022text} &34M&96.0&56.8&55.9&63.1&43.3&42.0&27.6&53.7&74.8&57.0\cr
				E5-base~\shortcite{wang2022text} &110M&95.4&65.6&59.4&66.3&47.3&\textbf{44.0}&32.0&\underline{69.3}&\underline{76.1}&61.1\cr
				\midrule
				PEARL-small&40M&\textbf{97.2}$_{\pm \text{0.1}}$&\underline{69.2$_{\pm \text{0.7}}$}&56.1$_{\pm \text{0.1}}$&\underline{69.7$_{\pm \text{0.1}}$}&\underline{48.1$_{\pm \text{0.1}}$}&43.4$_{\pm \text{0.2}}$&\textbf{48.7$_{\pm \text{0.7}}$}&61.0$_{\pm \text{1.1}}$&74.6$_{\pm \text{0.1}}$&\underline{63.1$_{\pm \text{0.2}}$}\cr
				PEARL-base&116M&\underline{97.1$_{\pm \text{0.0}}$}&\textbf{72.7$_{\pm \text{0.4}}$}&\textbf{60.9$_{\pm \text{0.3}}$}&\textbf{72.3$_{\pm \text{0.3}}$}&\textbf{50.2$_{\pm \text{0.2}}$}&\underline{43.6$_{\pm \text{0.4}}$}&\underline{40.9$_{\pm \text{0.2}}$}&\textbf{69.5$_{\pm \text{0.6}}$}&\textbf{76.3$_{\pm \text{0.0}}$}&\textbf{64.8$_{\pm \text{0.2}}$}\cr
				\bottomrule
			\end{tabular}
			\caption{Evaluations of various phrase-level tasks. For the AutoFJ, we report the average accuracy across 50 datasets. The best results are shown in bold and the second best results are underlined. Since the baseline String Distance cannot produce phrase embeddings, we only report its results on the AutoFJ as a reference.}%
			\label{tab:oevrall_result}%
			\vspace{-5pt}
		\end{threeparttable} 
	}
\end{table*}

\begin{table}[bt] 
	\centering 
	\tiny
	\setlength{\tabcolsep}{1.0mm}{
		\begin{threeparttable} 
			\begin{tabular}{ccccccccc}  
				\toprule
				\textbf{Model}
				&\textbf{Size}&\multicolumn{2}{c}{\textbf{\underline{Sentiment Analysis}}}&\multicolumn{2}{c}{\textbf{\underline{Intent Classification}}}&Avg\cr
				&&Twitter-S&Twitter-L&ATIS-S&ATIS-L&\cr
				Length&&(4.5)&(9.2)&(2.7)&(12.1)\cr
				\midrule
				SimCSE~\shortcite{gao2021simcse} &110M&70.4$_{\pm \text{0.3}}$&74.5$_{\pm \text{0.2}}$&91.2$_{\pm \text{0.5}}$&96.8$_{\pm \text{0.1}}$&83.2\cr
				Phrase-BERT~\shortcite{wang2021phrase} &110M&71.9$_{\pm \text{0.1}}$&77.0$_{\pm \text{0.2}}$&50.6$_{\pm \text{1.4}}$&79.5$_{\pm \text{2.7}}$&69.8\cr
				UCTopic~\shortcite{li2022uctopic} &240M&60.3$_{\pm \text{0.1}}$&70.6$_{\pm \text{0.3}}$&26.9$_{\pm \text{0.0}}$&72.2$_{\pm \text{0.0}}$&57.5\cr
				E5-small~\shortcite{wang2022text} &34M&70.7$_{\pm \text{0.4}}$&78.1$_{\pm \text{1.2}}$&92.7$_{\pm \text{0.0}}$&94.1$_{\pm \text{0.1}}$&83.9\cr
				E5-base~\shortcite{wang2022text} &110M&72.4$_{\pm \text{0.2}}$&\textbf{79.5}$_{\pm \text{0.4}}$&93.0$_{\pm \text{0.6}}$&96.2$_{\pm \text{0.3}}$&\underline{85.3}\cr
				\midrule
				PEARL-small&40M&\underline{72.8$_{\pm \text{0.2}}$}&\underline{78.5$_{\pm \text{0.5}}$}&\textbf{93.7}$_{\pm \text{0.5}}$&\underline{96.7$_{\pm \text{0.1}}$}&\textbf{85.4}\cr
				PEARL-base&116M&\textbf{73.7$_{\pm \text{0.3}}$}&77.1$_{\pm \text{0.1}}$&\underline{93.2$_{\pm \text{0.7}}$}&\textbf{97.4$_{\pm \text{0.1}}$}&\textbf{85.4}\cr
				\bottomrule
			\end{tabular}
			\caption{Evaluations of text classification tasks. We run each model 10 times and report the average accuracy. ``S'' and ``L'' mean short and long, respectively. The best results are shown in bold and the second best results are underlined.}%
			\label{tab:short_text_result}%
			\vspace{-5pt}
		\end{threeparttable} 
	}
\end{table}

For short text datasets, we consider two tasks: 
(1) \textbf{Sentiment Analysis}. We use a Twitter corpus~\footnote{\url{https://huggingface.co/datasets/carblacac/twitter-sentiment-analysis}} for this goal due to its short length. Two datasets are constructed based on this corpus: Twitter-S and Twitter-L, which contain 10,000 short Twitter sentences and 20,000 long Twitter sentences, respectively.  (2) \textbf{Intent Classification}. We use the ATIS (Airline Travel Information Systems) dataset~\cite{hemphill1990atis}, which consists of 5400 queries with 8 intent categories. We constructed two subsets, ATIS-S and ATIS-L, based on the length of query sentences.

\subsection{Implementation Details}\label{sec:implementation}

All approaches are implemented with PyTorch \cite{paszke2019pytorch} and HuggingFace~\cite{wolf2020transformers}. 
We use three NVIDIA Tesla V100S PCIe 32 GB for all experiments.
We test two versions of PEARL, PEARL-small and PEARL-base, initialized by E5-small and E5-base~\cite{wang2022text}, respectively. We then fine-tune them on our constructed phrase dataset for two epochs. 
The hyperparameters are selected by using grid search (see Figure~\ref{fig:hp_bird}).
The batch size is 512 (the maximum capacity for a single GPU), and we use Adam~\cite{kingma2014adam} with a learning rate of $3e-5$ for optimization. The learning rate is exponentially decayed for every 2000 steps with a rate of 0.98.
The temperature $\tau$ is the default value of 0.07 and the number of hard negatives is 2 for each mini-batch.
Each data augmentation method is randomly used during training.
We fine-tune PEARL three times with different seeds and report the average score.

\begin{table*}[!t]  
	\centering  
	\scriptsize
	\begin{threeparttable} 
		\begin{tabular}{lcccccc}  
			\toprule  
			\multirow{1}{*}{Model}&Paraphrase Classification&Phrase Similarity&Entity Retrieval&Entity Clustering&Fuzzy Join&Avg\cr    
			\midrule
			PEARL-small&83.2$_{\pm \text{0.4}}$&62.9$_{\pm \text{0.1}}$&45.8$_{\pm \text{0.2}}$&54.9$_{\pm \text{0.9}}$&74.6$_{\pm \text{0.1}}$&63.1$_{\pm \text{0.2}}$\cr 
			\midrule
			- Phrase DA &\textbf{82.6}$_{\pm \text{0.1}}$ \textcolor{brandeisblue}{$\downarrow$}&\textbf{61.2}$_{\pm \text{0.4}}$ \textcolor{brandeisblue}{$\downarrow$}&\textbf{41.0}$_{\pm \text{0.6}}$ \textcolor{brandeisblue}{$\downarrow$}&52.1$_{\pm \text{1.7}}$ \textcolor{brandeisblue}{$\downarrow$}&\textbf{72.7}$_{\pm \text{0.3}}$ \textcolor{brandeisblue}{$\downarrow$}&\textbf{60.7}$_{\pm \text{0.3}}$ \textcolor{brandeisblue}{$\downarrow$}\cr
			- Entity Type  &82.9$_{\pm \text{1.0}}$ \textcolor{brandeisblue}{$\downarrow$}&63.7$_{\pm \text{0.2}}$ 
			\textcolor{carrotorange}{$\uparrow$}&44.3$_{\pm \text{0.2}}$ \textcolor{brandeisblue}{$\downarrow$}&\textbf{45.7}$_{\pm \text{0.5}}$ \textcolor{brandeisblue}{$\downarrow$}&74.9$_{\pm \text{0.1}}$ \textcolor{carrotorange}{$\uparrow$}&60.9$_{\pm \text{0.1}}$ \textcolor{brandeisblue}{$\downarrow$}\cr	
			- Token DA  &82.7$_{\pm \text{0.4}}$ \textcolor{brandeisblue}{$\downarrow$}&62.8$_{\pm \text{0.4}}$ 
			\textcolor{brandeisblue}{$\downarrow$}&44.6$_{\pm \text{0.7}}$ \textcolor{brandeisblue}{$\downarrow$}&51.4$_{\pm \text{2.0}}$ \textcolor{brandeisblue}{$\downarrow$}&73.9$_{\pm \text{0.3}}$ \textcolor{brandeisblue}{$\downarrow$}&61.9$_{\pm \text{0.5}}$ \textcolor{brandeisblue}{$\downarrow$}\cr	
			- Hard Negatives &83.2$_{\pm \text{0.4}}$ \textcolor{x11gray}{$\updownarrow$}&63.2$_{\pm \text{0.4}}$ \textcolor{carrotorange}{$\uparrow$}&45.1$_{\pm \text{0.7}}$ \textcolor{brandeisblue}{$\downarrow$}&52.0$_{\pm \text{0.4}}$ \textcolor{brandeisblue}{$\downarrow$}&73.9$_{\pm \text{0.2}}$ \textcolor{brandeisblue}{$\downarrow$}&62.3$_{\pm \text{0.2}}$ \textcolor{brandeisblue}{$\downarrow$}\cr
			- Character Encoder  &82.8$_{\pm \text{0.4}}$ \textcolor{brandeisblue}{$\downarrow$}&63.3$_{\pm \text{0.4}}$ \textcolor{carrotorange}{$\uparrow$}&45.8$_{\pm \text{0.4}}$ \textcolor{x11gray}{$\updownarrow$}&51.7$_{\pm \text{0.7}}$ \textcolor{brandeisblue}{$\downarrow$}&73.9$_{\pm \text{0.2}}$ \textcolor{brandeisblue}{$\downarrow$}&62.3$_{\pm \text{0.2}}$ \textcolor{brandeisblue}{$\downarrow$}\cr
			- Character DA  &82.9$_{\pm \text{0.3}}$ \textcolor{brandeisblue}{$\downarrow$}&63.5$_{\pm \text{0.3}}$ \textcolor{carrotorange}{$\uparrow$}&45.3$_{\pm \text{0.6}}$ \textcolor{brandeisblue}{$\downarrow$}&51.6$_{\pm \text{1.4}}$ \textcolor{brandeisblue}{$\downarrow$}&74.5$_{\pm \text{0.4}}$ \textcolor{brandeisblue}{$\downarrow$}&62.4$_{\pm \text{0.5}}$ \textcolor{brandeisblue}{$\downarrow$}\cr
			\bottomrule  
		\end{tabular}
		\caption{Ablation study. DA means Data Augmentation. The biggest drop is in bold.}\label{tab:ablation}
	\end{threeparttable}  
\end{table*}

\subsection{Competitors}

We compare our approach to the following competitors:
\textbf{String Distance} uses the Jaccard similarity of n-gram characters to compare two strings. 
\textbf{FastText}~\cite{bojanowski2017enriching} and \textbf{GloVe}~\cite{pennington2014glove} are two popular word embedding methods, and we average word embeddings in order to obtain phrasal representations. \textbf{Sentence-BERT}~\cite{reimers2019sentence} fine-tuned BERT on SNLI~\cite{bowman2015large} sentence pairs.  
\textbf{Phrase-BERT}~\cite{wang2021phrase} is a dedicated model for phrase representation fine-tuned on lexically diverse datasets. \textbf{UCTopic}~\cite{li2022uctopic} is an unsupervised contrastive learning framework for
context-aware phrase representations and topic mining. \textbf{E5}~\cite{wang2022text} is a powerful text embedding model that can transfer to a wide range of tasks.
E5 offers three model sizes: E5$_{small}$, E5$_{base}$, and E5$_{large}$, initialized from MiniLM~\cite{wang2021minilmv2}, BERT$_{base}$, and BERT$_{large}$. We do not compare to McPhrasy~\cite{cohen2022mcphrasy} because it is not publicly available.
\section{Results}

\subsection{Overall Performance}
Table~\ref{tab:oevrall_result} shows the experimental results across five phrase tasks.
We first note that PEARL-base achieves the best performance on average, obtaining the best score on 6 of 9 datasets.
Second, our framework brings significant improvements to the corresponding backbone language models.
Specifically, PEARL-base improves E5-base by 3.7 absolute percentage points on average and the corresponding improvement of PEARL-small is 6.1 absolute percentage points.
Moreover, PEARL-small with 40 million parameters outperforms other competitors, and this result validates our claim that a small model can obtain competitive results with a big model for short text representations. 

Apart from these phrase tasks, we conduct experiments on short text classification to show a practical usage of our PEARL model and the results are shown in Table~\ref{tab:short_text_result}.
While PEARL is able to outperform other phrase models like Phrase-BERT and UCTopic, there is no statistical difference compared to other sentence models like SimCSE (\textit{BERT-unsup}) and E5.
It is worth mentioning that our model brings a benefit on very short texts (Twitter-S and ATIS-S). 

We conclude that our PEARL framework can produce high-quality representations for phrases and short texts across various tasks. If the length of input texts is very short (e.g., less than six tokens), it is beneficial to use PEARL embeddings. 

\subsection{Ablation Study}
We vary components of PEARL to validate architectural choices. We use PEARL-small as the baseline.
We fine-tune each variation of PEARL-small in the same experimental setting and test it across five phrase tasks. 
All results are shown in Table~\ref{tab:ablation}.

\paragraph{Entity Type Classification.}
If entity type classification is removed, the average performance decreases by 2.2 percentage points and drops dramatically for the entity clustering task. 
This validates our claim that adding phrase type information enhances representation capabilities.

\begin{figure*}[t]%
	\centering
	\subfloat[\centering Phrase-BERT (110M) ]{{\includegraphics[width=0.25\textwidth]{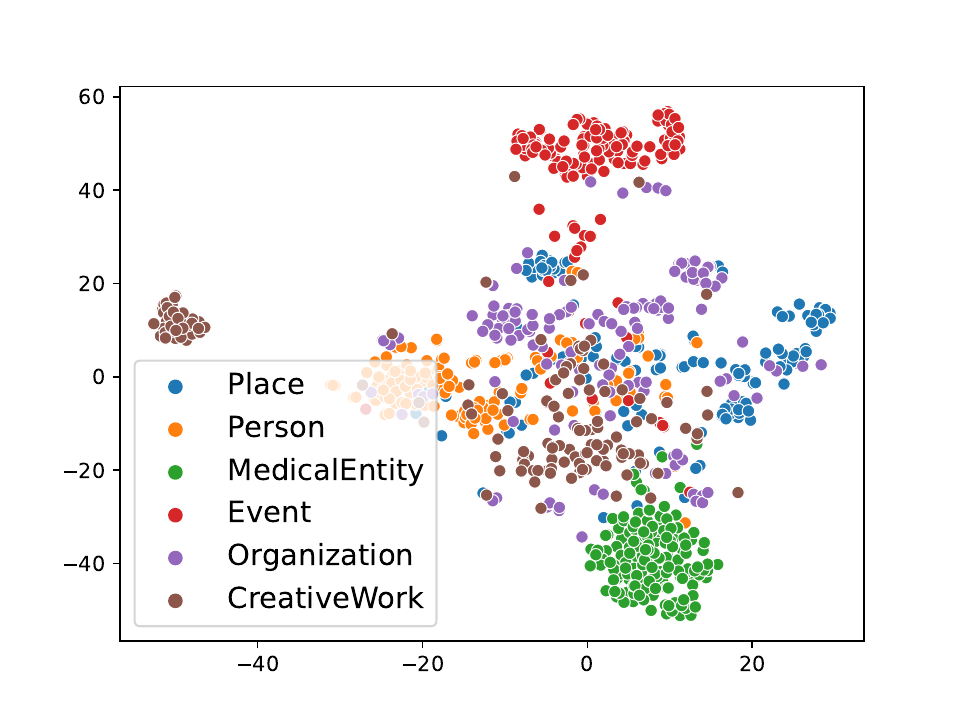} }}%
	\subfloat[\centering UCTopic (240M) ]{{\includegraphics[width=0.25\textwidth]{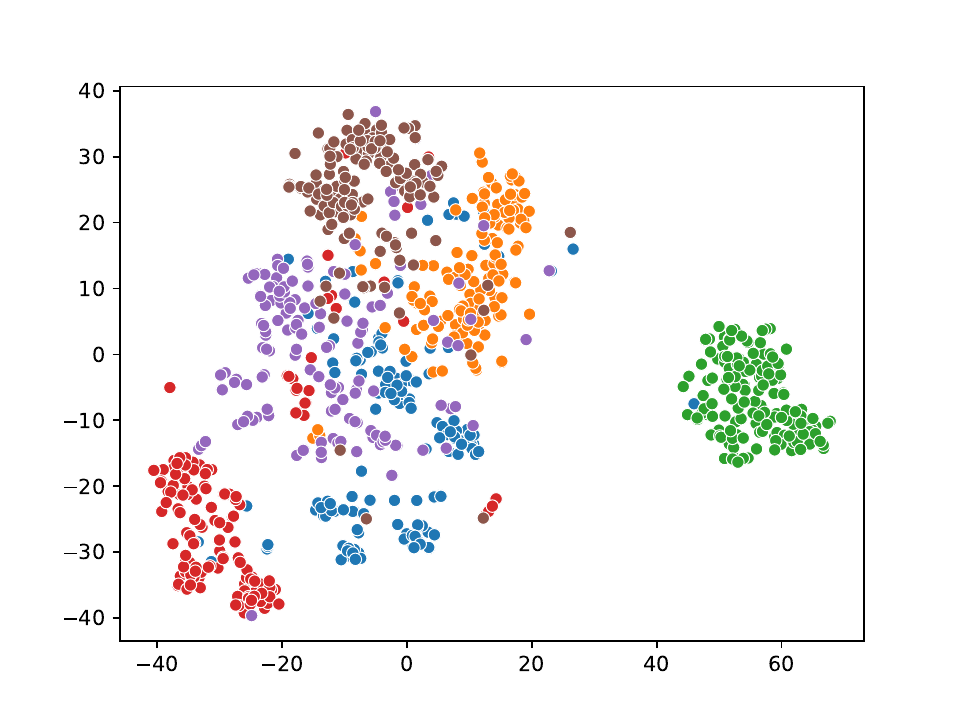} }}%
	\subfloat[\centering E5-base (110M) ]{{\includegraphics[width=0.25\textwidth]{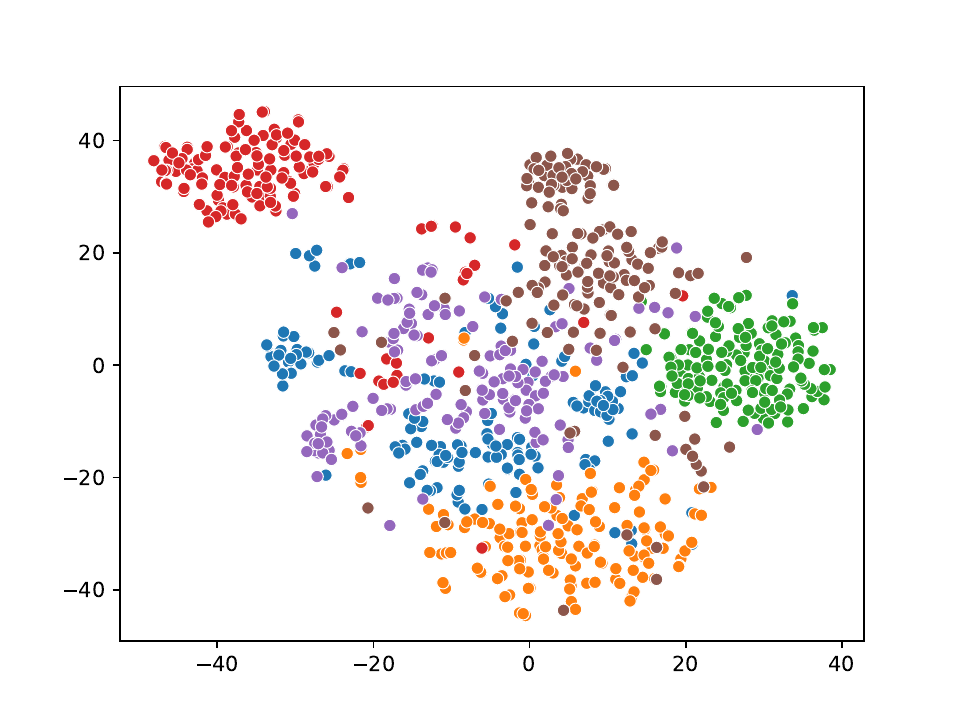} }}
	\subfloat[\centering PEARL-small (40M) ]{{\includegraphics[width=0.25\textwidth]{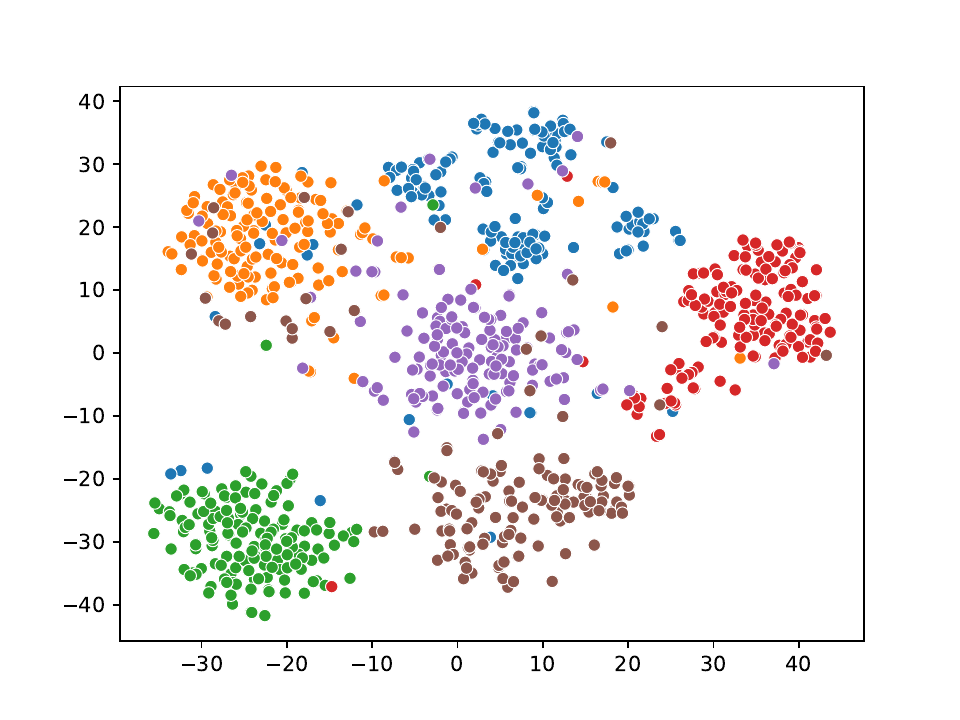} }}
	\caption{t-SNE visualizations of phrase embeddings generated by different models. We randomly selected 100 samples for each entity type from YAGO 4 (\texttt{Place, Person, MeidicalEntity, Event, Organization, CreativeWork}). Markers with the same color are supposed to be grouped together.   }
	\label{fig:type_visualization}%
	
\end{figure*}

\paragraph{Character Encoder.}
PEARL uses LOVE~\cite{chen2022imputing} to capture morphological variations of phrases. Removing LOVE causes a drop of 0.8 percentage points on average, especially for the entity-clustering task (-3.2). 

\paragraph{Data Augmentation.} 
PEARL uses data augmentation at three levels of granularity: character-level, token-level, and phrase-level methods.
To validate the effect of each level, we stop using a particular augmentation during fine-tuning. 
We find that character-level augmentation is beneficial mainly for the tasks of entity clustering (-3.3) and Entity Retrieval (-0.5).
Token-level augmentations create lexically diverse positive phrases, and removing these samples degrades performances across all five tasks.
Phrase-level augmentation has the strongest impact on the representation capabilities of a model.
Removing all augmentations results in an average drop of 2.4 percentage points.

\paragraph{Hard Negatives.} As random in-batch negatives contain relatively less information to learn, we insert a number of hard negatives into each batch. These negatives share similar surface forms with the original phrases but differ in their meanings.  We find that adding hard negatives brings decent improvements (+0.8 on average), especially considering the nearly zero additional cost of this strategy.

\begin{table}[!t]  
	\centering  
	\scriptsize
	\begin{threeparttable} 
		\begin{tabular}{cccccc}  
			\toprule  
			\multirow{1}{*}{Model}&BERT&RoBERTa&ALBERT&SpanBERT&LUKE\cr   
			\midrule
			Original&39.4&33.2&33.6&29.6&31.9\cr 
			+ PEARL &57.1&53.4&52.5&50.6&52.7\cr
			
			$\Delta$ &17.7 $\uparrow$&20.2 $\uparrow$&18.9 $\uparrow$&21.0 $\uparrow$&20.8 $\uparrow$\cr
			\bottomrule  
		\end{tabular}
		\caption{The performances of language models after using our framework. The results are the average score across five phrase tasks.}\label{tab:generalizability}
	\end{threeparttable}  
\end{table}

\subsection{Visualization}
To demonstrate more intuitively the improved quality of phrasal representations, we visualize embeddings generated by different models. 
Specifically, we use six types of entities from YAGO~4~\cite{pellissier2020yago} in this experiment: \texttt{Place, Person, MeidicalEntity, Event, Organization, CreativeWork}. 
For each type, 100 entity names are randomly sampled from the entire set and we feed them into the four models for obtaining phrase embeddings. Then, we apply t-SNE to reduce them to 2 dimensions for visualization. As Figure~\ref{fig:type_visualization} shows, PEARL can effectively cluster the same types of phrases together.

\subsection{Generalizability of Our Framework}
We now demonstrate that PEARL can enhance the phrase representations of various language models.
Beyond E5, we test five other language models: BERT~\cite{devlin2018bert}, RoBERTa~\cite{liu2019roberta}, ALBERT~\cite{lan2019albert}, SpanBERT~\cite{joshi2020spanbert},
and LUKE~\cite{yamada2020luke}.
We first check the original performance of each language model across five phrase tasks and then use PEARL to fine-tune them by following the same experimental setting as before (but using 30\% of training samples to save time). 
Table~\ref{tab:generalizability} shows that PEARL consistently obtains significant enhancements, showing that our method can be generalized to various models. 

\begin{figure}[t]%
	\vspace{-25pt}
	\centering
	\subfloat[\centering Learning Rate (e-5)]{{\includegraphics[width=0.25\textwidth]{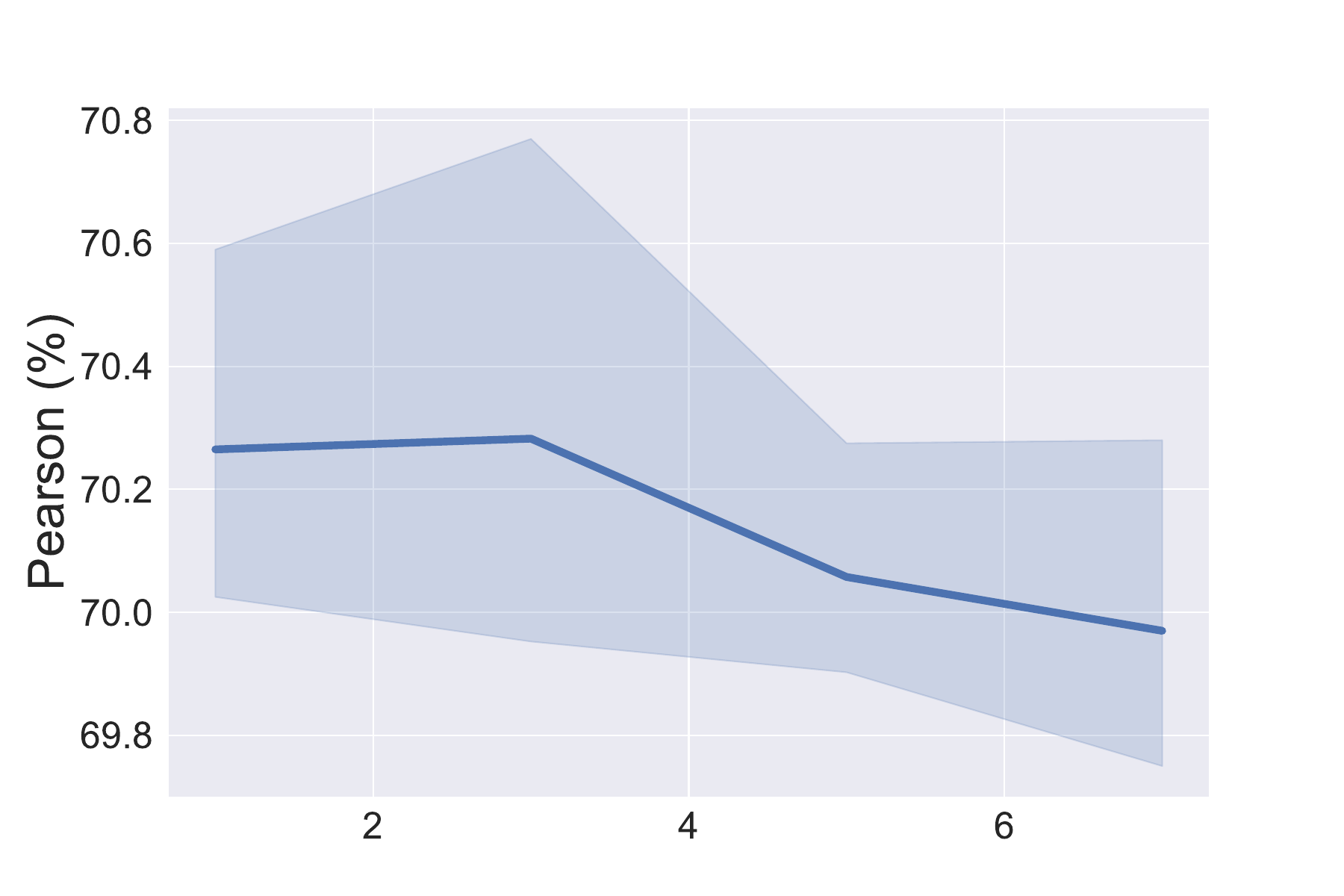} }}%
	\subfloat[\centering Hard Negatives ]{{\includegraphics[width=0.25\textwidth]{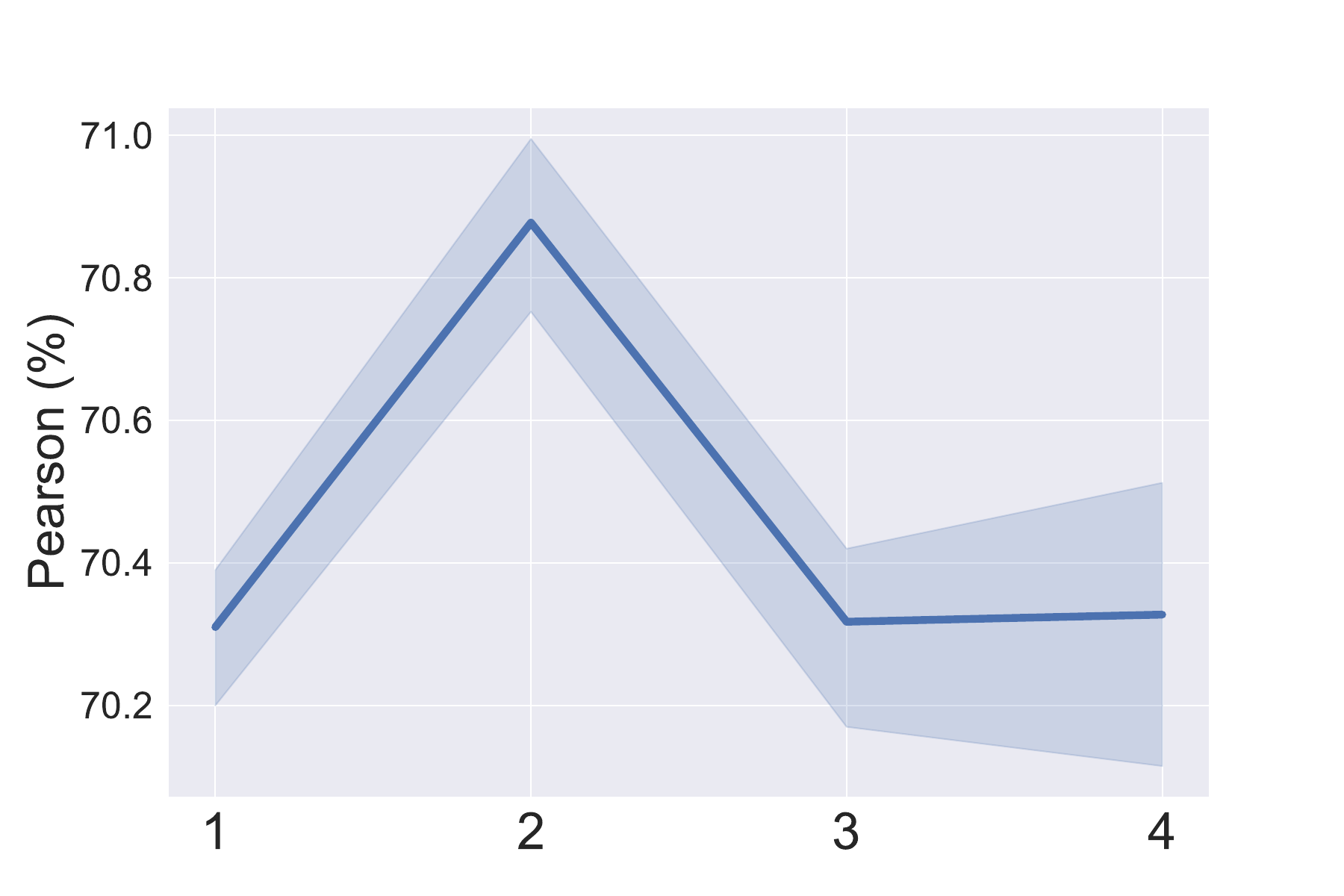} }}%
	\caption{Hyperparameter selection on BIRD dataset.}
	\label{fig:hp_bird}%
	
\end{figure}

\subsection{Hyperparameter Selection}
Figure~\ref{fig:hp_bird} shows the performances on the BIRD datasets by varying learning rates and numbers of hard negatives. We observe that a learning rate of 3e-5 and using 2 hard negatives in each batch can yield better phrase embeddings.

\section{Conclusion}
In this study, we have presented PEARL, a novel contrastive learning framework for more powerful phrase representations. 
PEARL incorporates phrase type information and morphological features, and thereby captures better the nuances of phrases.
Furthermore, PEARL enriches training samples with
distinct granularities of data augmentations. 
Our empirical results show that it improves phrase embeddings for a wide range of tasks, from paraphrase classification to entity retrieval, useful in applications across NLP and data engineering.
Adding character-level support to language models appears crucial to success on short texts. Indeed, these provide much less context than full paragraphs and thus it is important to go beyond the tokens of the original language model that mainly capture word stems.

\section*{Limitations}
One potential limitation is that our PEARL may not provide significant advantages when dealing with long sentences. Since PEARL is specifically dedicated to modeling morphological variations of short texts by using context-free input,  current PEARL models do not capture long-distance contextual semantics very well, which can limit their performances and benefits on long texts. 

\section*{Acknowledgements}
This work was partially funded by projects NoRDF (ANR-20-CHIA-0012-01) and LearnI (ANR-20-CHIA-0026).

\bibliography{custom}

\appendix

\setcounter{table}{0}   
\setcounter{figure}{0}
\renewcommand{\thetable}{A\arabic{table}}
\renewcommand{\thefigure}{A\arabic{figure}}
\setcounter{equation}{0}

\section{Appendix}
\label{sec:appendix}

\subsection{Details of Datasets}\label{sec:details_of_datasets}
\subsubsection{Phrase Datasets}
\paragraph{Paraphrase Classification (PC)}$\!\!\!$judges whether two phrases convey the same meaning. We use two paraphrase classification datasets used by Phrase-BERT~\cite{wang2021phrase}: \textbf{PPDB} and \textbf{PPDB-filtered}. 
PPDB is constructed from PPDB 2.0~\cite{pavlick2015ppdb}, which includes 23,364 phrase pairs by sampling examples from PPDB-small with a high score, and negative examples are randomly selected from the dataset.
PPDB-filtered contains more challenging samples, which are obtained by removing phrase pairs with lexical overlap cues. In total, there are 19,416 phrase pairs.
We follow the setting of previous work for experiments~\cite{wang2021phrase}, where a simple classifier layer (multilayer perceptron with a ReLu activation) is added on top of the concatenated embeddings
of a phrase pair. We measure accuracy.

\paragraph{Phrase Similarity (PS)}$\!\!\!$aims to calculate the semantic similarity for phrase pairs. We use two datasets, \textbf{Turney}~\cite{turney2012domain} and \textbf{BIRD}~\cite{asaadi2019big}.
Turney evaluates bigram compositionality. A model is supposed to select the most similar unigram from five candidates given a bigram input. The dataset has 2180 samples and the metric is accuracy.
BIRD is a fine-grained and human-annotated bigram relatedness dataset, which contains 3345 English term pairs. 
Each pair of phrases has a relatedness score between 0 and 1, and the metric for this dataset is the Pearson correlation coefficient. 

\paragraph{Entity Retrieval (ER)}$\!\!\!\!$aims to retrieve a standard entity from a reference knowledge base given a textual mention of that entity. We consider a particularly challenging form of the task, where the mention is given without any context, and the reference knowledge base provides only the canonical name of the entity. For example, given the mention \textit{``NYTimes''}, the goal is to determine the canonical entity \textit{``The New York Times''} in Wikidata. We construct two entity retrieval datasets by using a general knowledge base \textbf{Yago}~\cite{pellissier2020yago} and a biomedical terminology \textbf{UMLS}~\cite{bodenreider2004unified}, respectively. 
Both Yago and UMLS offer alternate names for an entity, and we randomly selected 10K of these alternate names as mentioned. The canonical names of the entities serve as the reference dictionary and there are no duplicate names in the dictionary.
The dictionary size of Yago and UMLS is 572K and 750K, respectively.
To accelerate the inference, we use Faiss~\cite{johnson2019billion} with all competing systems to do an approximate search. 
The metric here is top-1 accuracy.

\paragraph{Entity Clustering (EC)}$\!\!\!$tests whether the phrase embeddings can be grouped together according to their semantic categories. 
We use the general-purpose \textbf{CoNLL~03}~\cite{sang2003introduction} benchmark and the biomedical \textbf{BC5CDR}~\cite{li2016biocreative} benchmark. 
CoNLL~03 consists of 3,453 sentences with entities, and the three entity types are used in the experiment: \texttt{Person, Location}, and \texttt{Organization}.
BC5CDR has 7,095 sentences with two types of entities: \texttt{Disease} and \texttt{Chemical}. 
We apply KMeans~\cite{macqueen1967some} to the embeddings generated by a phrase representation model and use the NMI (normalized mutual information) metric.

\paragraph{Fuzzy Join (FJ)}$\!\!\!$is an important database operator widely used
in practice (also known as fuzzy-match), which matches record pairs from two tables. We use the \textbf{AutoFJ} benchmark~\cite{li2021auto}, which contains 50 diverse fuzzy-join datasets derived from DBpedia~\cite{lehmann2015dbpedia}. It aims to match entity names that have changed over time (e.g., \textit{``2012 Wisconsin Badgers football team''} and \textit{``2012 Wisconsin Badgers football season''}). In this experiment, we use the left table names as reference tables and the right table names as input tables. We report the average accuracy across all datasets.  

All experiments except paraphrase classification are conducted without fine-tuning.

\begin{figure}[t]%
	{{\includegraphics[width=0.50\textwidth]{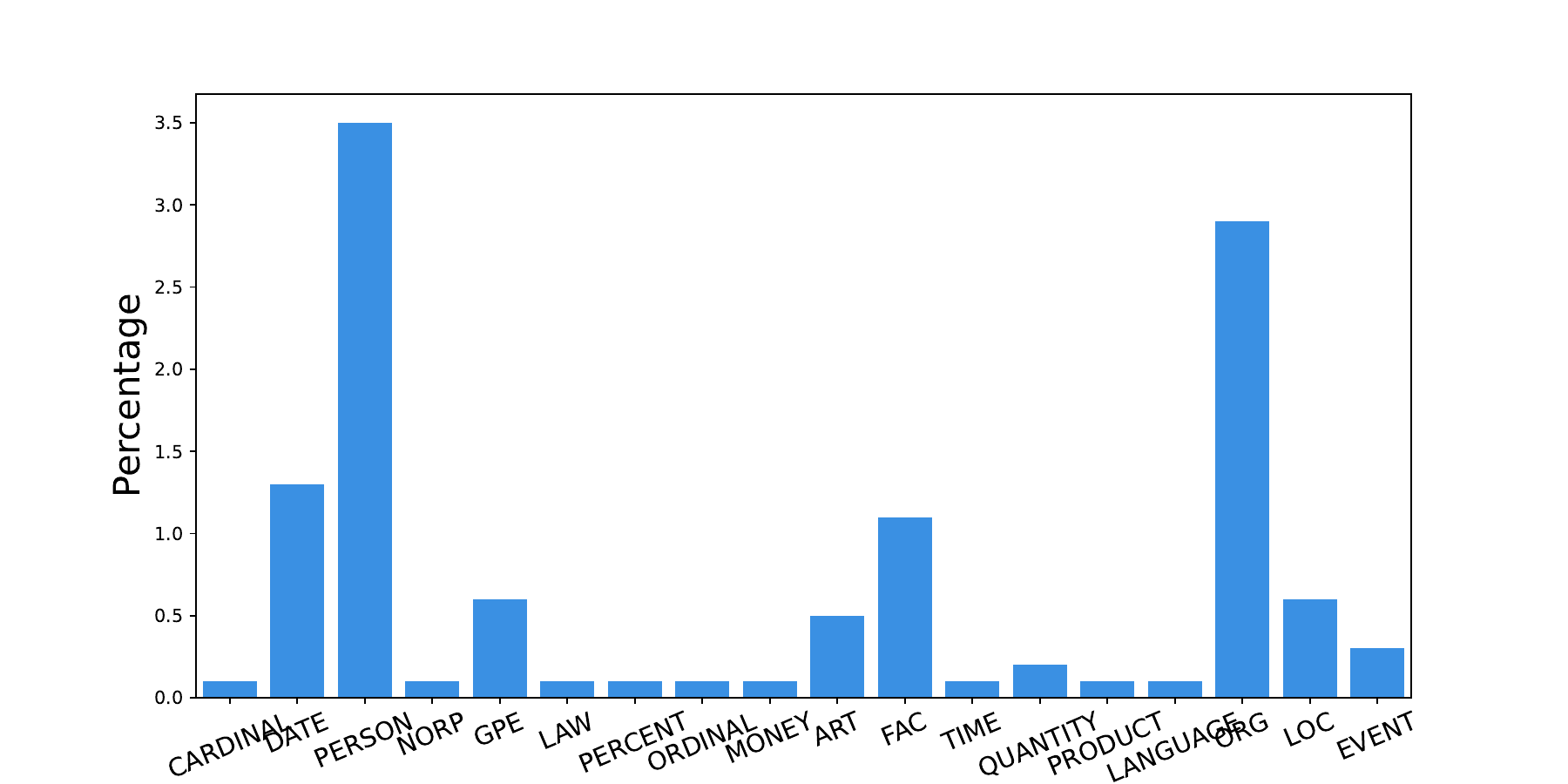} }}%
	\caption{Distributions of each entity type (without the \texttt{OTHER} tag, with 88.5\%).}
	\label{fig:phrase_type_dis}%
	\vspace{-10pt}
\end{figure}

\subsubsection{Short Text Datasets}
\paragraph{Sentiment Analysis (SA)} analyzes texts to determine whether the emotion is positive or negative. We use a Twitter corpus for this goal due to its short length. Two datasets are constructed based on this corpus: Twitter-S and Twitter-L, which contain 10,000 short Twitter sentences and 20,000 long Twitter sentences, respectively. 

\paragraph{Intent Classification (IC)} identifies customer's intents from text queries. 
We use ATIS (Airline Travel Information Systems) dataset~\cite{hemphill1990atis}, which consists of 5400 queries with 8 intent categories. We constructed two subsets, ATIS-S and ATIS-L, based on the length of query sentences.

For the two short text classification tasks, we add a classifier layer (multilayer perceptron with a ReLu activation) on top of the text embeddings and report the average accuracy across 10 times run.

\end{document}